\documentclass[10pt,twocolumn,letterpaper]{article}

\usepackage{wacv}
\usepackage{times}
\usepackage{epsfig}
\usepackage{graphicx}
\usepackage{amsmath}
\usepackage{amssymb}

\usepackage{bbm}
\usepackage{subcaption}
\usepackage{multirow}
\usepackage{hhline}
\usepackage{color,soul}
\usepackage{array}
\usepackage[update,prepend]{epstopdf}

\newcolumntype{C}[1]{>{\centering}m{#1}}

\wacvfinalcopy 

\def\wacvPaperID{105} 

\ifwacvfinal\fi
\begin{document}

\title{Fashion Apparel Detection: The Role of Deep Convolutional Neural Network and Pose-dependent Priors}

\author{Kota Hara\thanks{This work was done while the first author was an intern at eBay Research} \\
University of Maryland, College Park\\
{\tt\small kotahara@umd.edu}
\and
Vignesh Jagadeesh \hspace{0.5cm} Robinson Piramuthu \\
eBay Research\\
{\tt\small \{vjagadeesh,  rpiramuthu\}@ebay.com}
}

\maketitle
\ifwacvfinal\thispagestyle{empty}\fi

\begin{abstract}
In this work, we propose and address a new computer vision task, which we call fashion item detection, where the aim is to detect various fashion items a person in the image is wearing or carrying. The types of fashion items we consider in this work include hat, glasses, bag, pants, shoes and so on. The detection of fashion items can be an important first step of various e-commerce applications for fashion industry. Our method is based on state-of-the-art object detection method pipeline which combines object proposal methods with a Deep Convolutional Neural Network. Since the locations of fashion items are in strong correlation with the locations of body joints positions, we incorporate contextual information from body poses in order to improve the detection performance. Through the experiments, we demonstrate the effectiveness of the proposed method.
\end{abstract}

\section{Introduction}
\label{sec:introduction}
In this work, we propose a method to detect fashion apparels a person in an image is wearing or holding. The types of fashion apparels include hat, bag, skirt, etc. Fashion apparel spotting has gained considerable research traction in the past couple of years. A major reason is due to a variety of applications that a reliable fashion item spotter can enable. For instance, spotted fashion items can be used to retrieve similar or identical fashion items from an online inventory. 



Unlike most prior works on fashion apparel spotting which address the task as a specialization of the semantic segmentation to the fashion domain, we address the problem as an object detection task where the detection results are given in the form of bounding boxes. Detection-based spotters are more suitable as (a) bounding boxes suffice to construct queries for the subsequent visual search, (b) it is generally faster and have lower memory footprint than semantic segmentation, (c) large scale pixel-accurate training data is extremely hard to obtain, while it is much easier to get training data as bounding boxes, and (d) detection is done at instance-level while semantic segmentation does not differentiate multiple instances belonging to the same class.  To the best of our knowledge, our work is the first detection-based (as opposed to segmentation-based) fashion item spotting method.

Although any existing object detection methods can be possibly applied, the fashion apparel detection task poses its own challenges such as (a) deformation of clothing is large, (b) some fashion items classes are extremely similar to each other in appearance (e.g., skirt and bottom of short dress), (c) the definition of fashion item classes can be ambiguous (e.g., pants and tights), and (d) some fashion items are very small (e.g., belt, jewelry). In this work, we address some of these challenges by incorporating state-of-the-art object detectors with various domain specific priors such as pose, object shape and size.

The state-of-the-art object detector we employ in this work is R-CNN~\cite{Girshick2014}, which combines object proposals with a Convolutional Neural Network~\cite{Fukushima1980,Lecun1998}. The R-CNN starts by generating a set of object proposals in the form of bounding boxes. Then image patches are extracted from the generated bounding boxes and resized to a fixed size. The Convolutional Neural Network pretrained on a large image database for the image classification task is used to extract features from each image patch. SVM classifiers are then applied to each image patch to determine if the patch belongs to a particular class. The R-CNN is suitable for our task as it can detect objects with various aspect ratios and scales without running a scanning-window search, reducing the computational complexity as well as false positives. 

It is evident that there are rich priors that can be exploited in the fashion domain. For instance, handbag is more likely to appear around the wrist or hand of the person holding them, while shoes typically occur near feet. The size of items are typically proportional to the size of a person. Belts are generally elongated. One of our contributions is to integrate these domain-specific priors with the object proposal based detection method. These priors are learned automatically from the training data.


We evaluate the detection performance of our algorithm on the previously introduced Fashionista dataset~\cite{Yamaguchi2012} using a newly created set of bounding box annotations. We convert the segmentation results of state-of-the-art fashion item spotter into bounding box results and compare with the results of the proposed method. The experiments demonstrate that our detection-based approach outperforms the state-of-the art segmentation-based approaches in mean Average Precision criteria.




The rest of the paper is organized as follows. Section~\ref{sec:related_work} summarizes related work in fashion item localization. Our proposed method is detailed in Section~\ref{sec:proposed_method} where we start with object proposal, followed by classification of these proposals using a combination of generative and discriminative approaches. Section~\ref{sec:experiments} validates our approach on the popular Fashionista Dataset~\cite{Yamaguchi2012} by providing both qualitative and quantitative evaluations. Finally, Section~\ref{sec:conclusions} contains closing remarks.

\begin{figure}[t]
  \centering
    \includegraphics[width=0.46\textwidth]{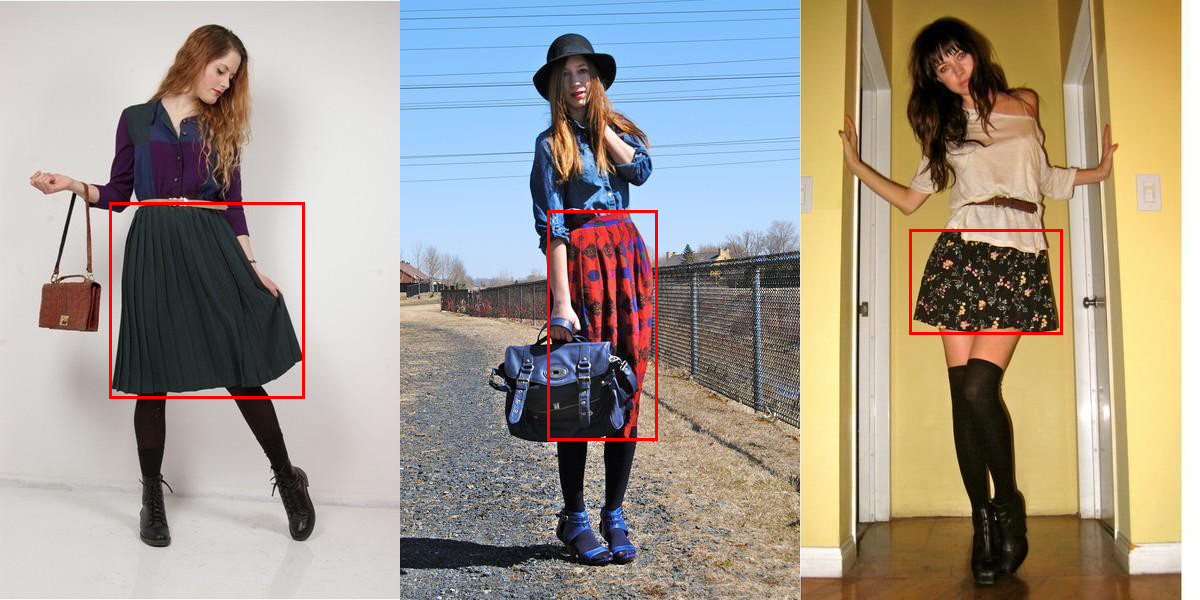}
  \caption{Bounding boxes of three different instances of ``skirt'' class. The aspect ratios vary significantly even though they are from the same object class.}
  \label{fig:example_skirt}
\end{figure}

\section{Related Work} 
\label{sec:related_work}
The first segmentation-based fashion spotting algorithm for general fashion items was proposed by \cite{Yamaguchi2012} where they introduce the Fashionista Dataset and utilize a combination of local features and pose estimation to perform semantic segmentation of a fashion image. In \cite{Yamaguchi2013}, the same authors followed up this work by augmenting the existing approach with data driven model learning, where a model for semantic segmentation was learned only from nearest neighbor images from an external database. Further, this work utilizes textual content along with image information. The follow up work reported considerably better performance than the initial work. We report numbers by comparing to the results accompanying these two papers. 

Apart from the above two works,~\cite{Hasan2010} also proposed a segmentation-based approach aimed at assigning a unique label from ``Shirt'', ``Jacket'', ``Tie'' and ``Face and skin'' classes to each pixel in the image. Their method is focused on people wearing suits.

There exist several clothing segmentation methods \cite{Gallagher2008,Hu2008a,Wang2011h} whose main goal is to segment out the clothing area in the image and types of clothing are not dealt with. In~\cite{Gallagher2008}, a clothing segmentation method based on graph-cut was proposed for the purpose of identity recognition. In~\cite{Hu2008a}, similarly to~\cite{Gallagher2008}, a graph-cut based method was proposed to segment out upper body clothing. \cite{Wang2011h} presented a method for clothing segmentation of multiple people. They propose to model and utilize the blocking relationship among people.

Several works exist for classifying types of upper body clothing~\cite{Bossard2012,Shen2014,Chen2012c}. In~\cite{Shen2014}, a structured learning technique for simultaneous human pose estimation and garment attribute classification is proposed. The focus of this work is on detecting attributes associated with the upper body clothing, such as collar types, color, types of sleeves, etc. Similarly, an approach for detecting apparel types and attributes associated with the upper bodies was proposed in~\cite{Bossard2012,Chen2012c}. Since localization of upper body clothing is essentially solved by upper body detectors and detecting upper body is relatively easy, the focus of the above methods are mainly on the subsequent classification stage. On the other hand, we focus on a variety of fashion items with various size which cannot be easily detected even with the perfect pose information. 

\cite{Yang2011a} proposed a real-time clothing recognition method in surveillance settings. They first obtain foreground segmentation and classify upper bodies and lower bodies separately into a fashion item class. In~\cite{Bourdev2011}, a poselet-based approach for human attribute classification is proposed. In their work, a set of poselet detectors are trained and for each poselet detection, attribute classification is done using SVM. The final results are then obtained by considering the dependencies between different attributes. In \cite{Wang2015WACV}, recognition of social styles of people in an image is addressed by Convolutional Neural Network applied to each person in the image as well as the entire image.

\begin{figure*}[tp]
  \centering
    \includegraphics[width=1.0\textwidth]{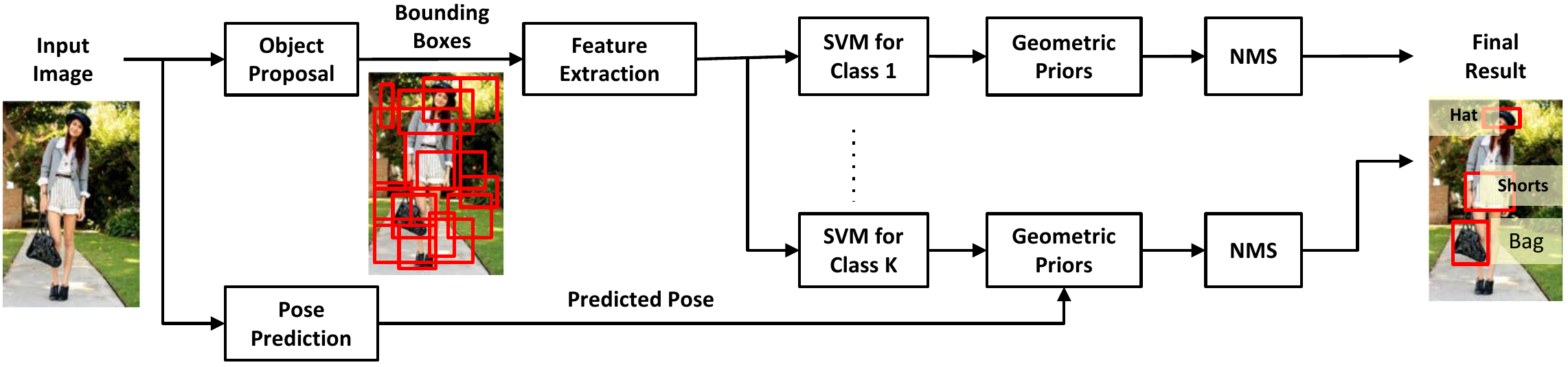}
  \caption{Overview of the proposed algorithm for testing stage. Object proposals are generated and features are extracted using Deep CNN from each object proposal. An array of 1-vs-rest SVMs are used to generate appearance-based posteriors for each class. Geometric priors are tailored based on pose estimation and used to modify the class probability. Non-maximum suppression is used to arbitrate overlapping detections with appreciable class probability.}
  \label{fig:overview}
\end{figure*}

\section{Proposed Method}
\label{sec:proposed_method}
The aim of the proposed method is to detect fashion items in a given image, worn or carried by a single person. The proposed method can be considered as an extension of the recently proposed R-CNN framework~\cite{Girshick2014}, where we utilize various priors on location, size and aspect ratios of fashion apparels, which we refer to as geometric priors. Specifically for location prior, we exploit strong correlations between pose of the person and location of fashion items. We refer to this as pose context. We combine these priors with an appearance-based posterior given by SVM to obtain the final posterior. Thus, the model we propose is a hybrid of discriminative and generative models.

The recognition pipeline of the proposed algorithm for the testing stage is shown in Figure~\ref{fig:overview}. Firstly, the pose of the person is estimated by an off-the-shelf pose estimator. Then, a set of candidate bounding boxes are generated by an object proposal algorithm. Image features are extracted  from the contents of each bounding box. These image features are then fed into a set of SVMs with a sigmoid function to obtain an appearance-based posterior for each class. By utilizing the geometric priors, a final posterior probability for each class is computed for each bounding box. The results are then filtered by a standard non-maximum suppression method~\cite{Felzenszwalb2010}. We explain the details of each component below.

\subsection{Object Proposal}
Object detection based on a sliding window strategy has been a standard approach~\cite{Felzenszwalb2010,Dalal2005,Viola2001,Bourdeva2009} where object detectors are exhaustively run on all possible locations and scales of the image. To accommodate the deformation of the objects, most recent works detect a single object by a set of part-specific detectors and allow the configurations of the parts to vary. Although a certain amount of deformation is accommodated, possible aspect ratios considered are still limited and the computation time increases linearly as the number of part detectors increases. 

In our task, the intra-class shape variation is large. For instance, as shown in Figure \ref{fig:example_skirt}, bounding boxes of three instances from the same ``skirt'' class have very different aspect ratios. Thus, for practical use, detection methods which can accommodate various deformations without significant increase in computation time are required.

In order to address these issues, we use object proposal algorithms \cite{Uijlings2013,Arbel2014} employed by state-of-the-art object detectors (i.e., R-CNN\cite{Girshick2014}). The object proposal algorithm generates a set of candidate bounding boxes with various aspect ratios and scales. Each bounding box is expected to contain a single object and the classifier is applied only at those candidate bounding boxes, reducing the number of false positives. For the classification step, an image patch within a bounding box is resized to a predefined size and image features are extracted. Since feature computation is done only at the generated bounding boxes, the computation time is significantly reduced while allowing various aspect ratios and scales. In this work, we employ Selective Search (SS) \cite{Uijlings2013} as the object proposal method.


\subsection{Image Features by CNN}
Our framework is general in terms of the choice of image features. However, recent results in the community indicate that features extracted by Convolutional Neural Network (CNN)~\cite{Fukushima1980,Lecun1998} with many layers perform significantly better than the traditional hand-crafted features such as HOG and LBP on various computer vision tasks \cite{Farabet2012,Krizhevsky2012,Sermanet2013,Zhang2014a}. However, in general, to train a good CNN, a large amount of training data is required. 

Several papers have shown that features extracted by CNN pre-trained on a large image dataset are also effective on other vision tasks. Specifically, a CNN trained on ImageNet database~\cite{Deng2009} is used for various related tasks as a feature extractor and achieve impressive performance~\cite{Donahue2014,Razavian2014}. In this work, we use~CaffeNet~\cite{jia2014} trained on ImageNet dataset as a feature extractor. We use a 4096 dimensional output vector from the second last layer (fc7) of CaffeNet as a feature vector.

\subsection{SVM training}
For each object class, we train a linear SVM to classify an image patch as positive or negative. The training patches are extracted from the training data with ground-truth bounding boxes. The detail of the procedure is described in Section \ref{sec:detectorTraining}.




\subsection{Probabilistic formulation}\label{sec:probablistic_formulation}
We formulate a probabilistic model to combine outputs from the SVM and the priors on the object location, size and aspect ratio (geometric priors) into the final posterior for each object proposal. The computed posterior is used as a score for each detection.

Let $B=(x_1,y_1,x_2,y_2)$ denote bounding box coordinates of an object proposal. Let $f$ denote image features extracted from $B$. We denote by $c = (l_x, l_y)$ the location of the bounding box center, where $l_x=(x_1+x_2)/2$ and $l_y=(y_1+y_2)/2$. We denote by $\mathrm{a}=\log ((y_2-y_1)/(x_2-x_1))$, the log aspect ratio of the bounding box and by $\mathrm{r}=\log ((y_2-y_1)+(x_2-x_1))$ the log of half the length of the perimeter of the bounding box. We refer to $c$, $\mathrm{a}$ and $\mathrm{r}$ as geometric features. 

Let $Y$ denote a set of fashion item classes and $y_{\mathrm{z}} \in \{+1, -1\}$ where $z \in Y$, denote a binary variable indicating whether or not $B$ contains an object belonging to $z$. Let $\mathbf{t}=(t_1,\dots, t_K) \in \mathrm{R}^{2 \times K}$ denote pose information, which is a set of $K$ 2D joint locations on the image. The pose information serves as additional contextual information for the detection.

We introduce a graphical model describing the relationship between the above variables and define a posterior of $y_z$ given $f$, $\mathbf{t}$, $c$, $\mathrm{a}$ and $\mathrm{r}$ as follows:
\begin{align}
p(y_z|f,c,\mathrm{a},\mathrm{r},\mathbf{t}) \propto p(y_z|f) p(c|y_z,\mathbf{t}) p(\mathrm{a}|y_z) p(\mathrm{r}|y_z,\mathbf{t})
\label{eq:graphicalmodel}
\end{align}
Here we assume that $p(\mathbf{t})$ and $p(f)$ are constant. The first term on the RHS defines the appearance-based posterior and the following terms are the priors on the geometric features. For each object proposal, we compute $p(y_z=1|f,c,\mathrm{a},\mathrm{r},\mathbf{t})$ and use it as a detection score. The introduced model can be seen as a hybrid of discriminative and generative models. In the following sections, we give the details of each component.

\subsection{Appearance-based Posterior}
We define an appearance based posterior $p(y_z=1|f)$ as
\begin{equation}
p(y_z=1|f) = \mathrm{Sig}( \mathrm{w}_{z}^{T} f ; \lambda_z )
\end{equation}
where $\mathrm{w}_z$ is an SVM weight vector for the class $z$ and $\lambda_z$ is a parameter of the sigmoid function $\mathrm{Sig}( \mathrm{x} ; \lambda_z) = 1/(1+\exp(-\lambda_z \mathrm{x}))$. The parameter $\lambda_z$ controls the shape of the sigmoid function. We empirically find that the value of $\lambda_z$ largely affects the performance. We optimize $\lambda_z$ based on the final detection performance on the validation set. 

\subsection{Geometric Priors}
\label{sec:geometric_priors}
\subsubsection*{Priors on Aspect Ratio and Perimeter}
The term $p(\mathrm{r}|y_z=1,\mathbf{t})$ is the prior on perimeter conditioned on the existence of an object from class $z$ and pose $\mathbf{t}$. Intuitively, the length of perimeter $\mathrm{r}$, which captures the object size, is useful for most of the items as there is a typical size for each item. Also, $\mathrm{r}$ is generally proportional to the size of a person. The size of the person can be defined using $\mathbf{t}$ in various ways. However, in this work, since the images in the dataset we use for experiments are already normalized such that the size of the person is roughly same, we assume $p(\mathrm{r}|y_z=1,\mathbf{t})=p(\mathrm{r}|y_z=1)$. 

The term $p(\mathrm{a}|y_z=1)$ is the prior on the aspect ratio of object bounding box conditioned on the existence of an object from class $z$. Intuitively, the aspect ratio $\mathrm{a}$ is useful for detecting items which have a distinct aspect ratio. For instance, the width of waist belt and glasses are most likely larger than their height. To model both $p(\mathrm{a}|y_z=1)$ and $p(\mathrm{r}|y_z=1)$, we use a 1-D Gaussian fitted by standard maximum likelihood estimation.

{\renewcommand{\arraystretch}{1.2}%
\begin{table*}[t]
\small
\centering
    \begin{tabular}{|l|C{5.5cm}|C{1.7cm}|C{3cm}||l|}
    \hline
    \textbf{New Class} & \textbf{Original Classes} & \textbf{Average Size in Pixel} & \textbf{Average Occurrence per Image}& \textbf{First and Second Key Joints} \\ \hline
    \textbf{Bag} & Bag, Purse, Wallet & 5,644 & 0.45 & Left hip, Right hip \\ \hline
    \textbf{Belt} & Belt & 1,068 & 0.23 & Right hip, Left hip \\ \hline
    \textbf{Glasses} & Glasses, Sunglasses & 541 & 0.16 & Head, Neck \\ \hline
    \textbf{Hat} & Hat & 2,630 & 0.14 & Neck, Right shoulder\\ \hline
    \textbf{Pants} & Pants, Jeans & 16,201 & 0.24 & Right hip, Left hip \\ \hline
    \textbf{Left Shoe} & \multirow{2}{5.5cm}{Shoes, Boots, Heels, Wedges, Flats, Loafers, Clogs, Sneakers, Sandals, Pumps} & 3,261 & 0.95 & Left ankle, Left knee \\ \hhline{-~---}
	\textbf{Right Shoe} &  & 2,827 & 0.93 & Right ankle, Right knee \\ \hline
    \textbf{Shorts} & Shorts & 6,138 & 0.16 & Right hip, Left hip \\ \hline
    \textbf{Skirt} & Skirt & 14,232 & 0.18 & Left hip, Right hip \\ \hline
    \textbf{Tights} & Tights, Leggings, Stocking & 10,226 & 0.32 & Right knee, Left knee \\ \hline
    \end{tabular}
\caption{The definition of new classes, their average size and the average number of occurrence per image are shown. The top 2 key body joints for each item as selected by the proposed algorithm are also shown. See Section~\ref{sec:dataset} for details.}
\label{table:classDefinition}
\end{table*}
}

\begin{figure}[t]
\centering
\subcaptionbox{Bag - Neck}{\includegraphics[width=0.48\linewidth]{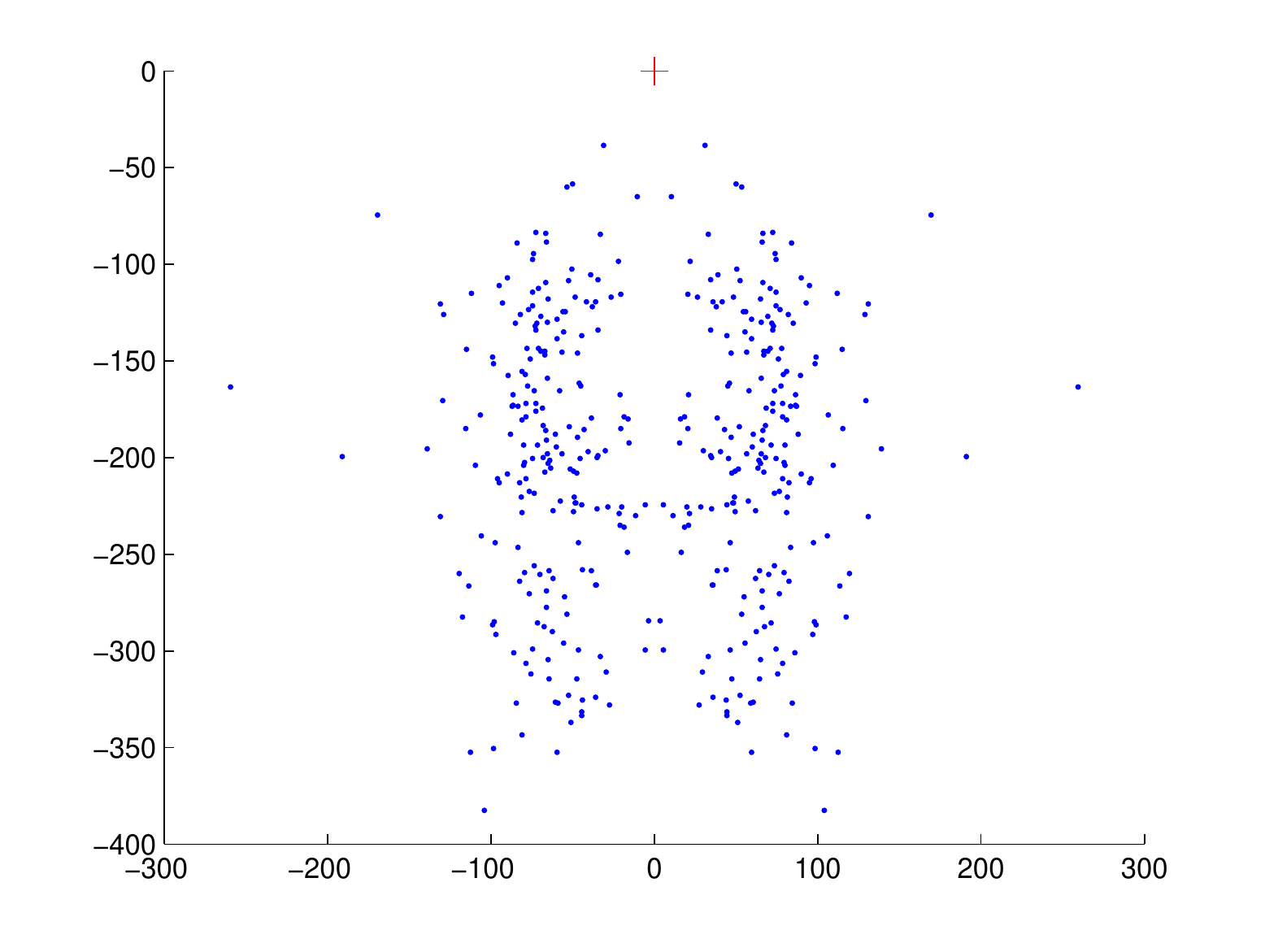}}
\subcaptionbox{Left Shoe - Left Ankle}{\includegraphics[width=0.48\linewidth]{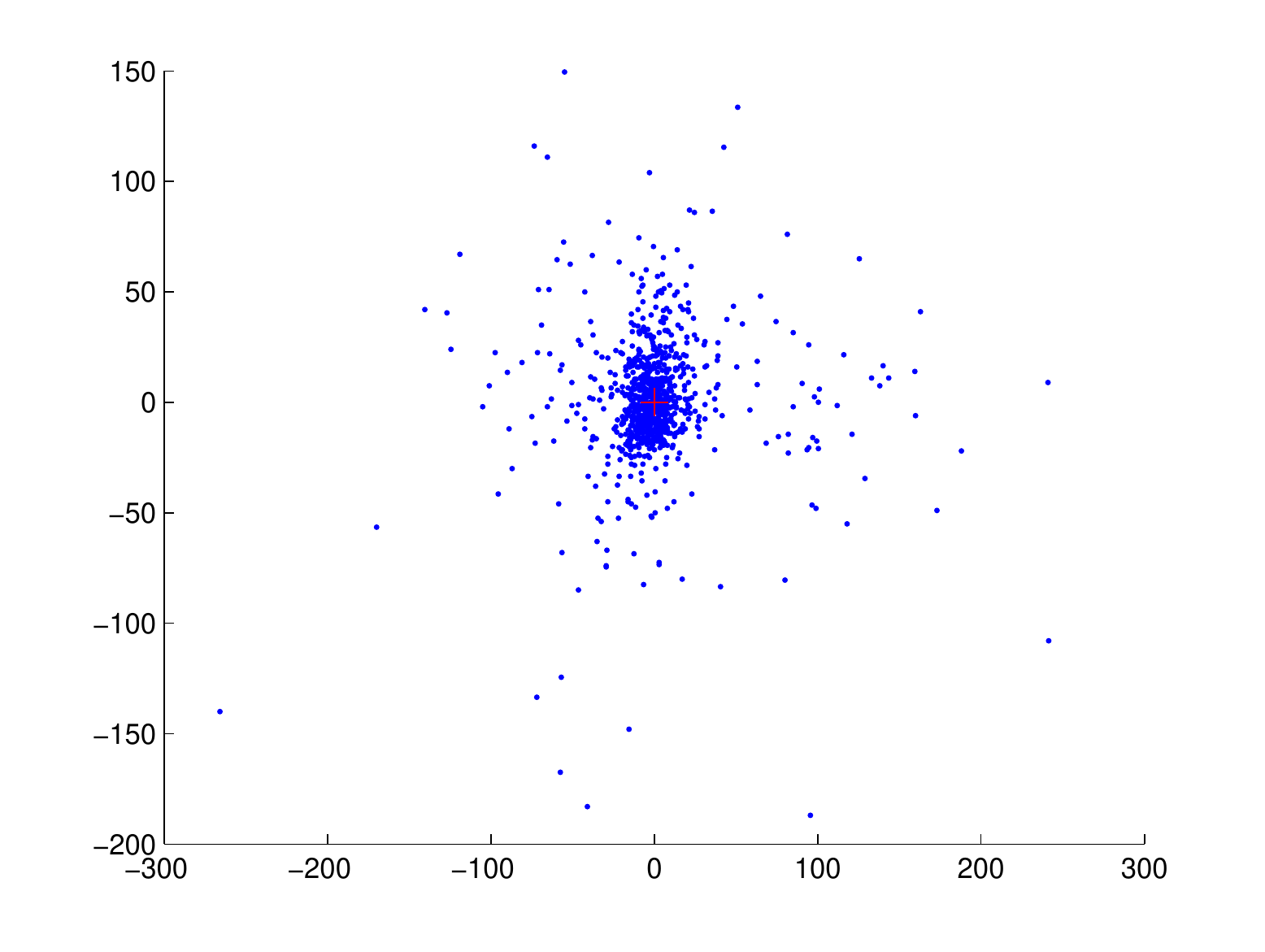}}
\caption{Distributions of relative location of item with respect to location of key joint. Key joint location is depicted as a red cross. (a) distribution of relative location of bag with respect to neck is multi-modal. (b) locations of left shoe and left ankle are strongly correlated and the distribution of their relative location has a single mode. See Section~\ref{sec:geometric_priors} for details.}
\label{fig:distribution}
\vspace{-1em}
\end{figure}

\subsubsection*{Pose dependent prior on the bounding box center}
We define a pose dependent prior on the bounding box center as
\begin{align}
p(c|y_z=1,\mathbf{t})
&=\Pi_{k\in T_z} p(l_x,l_y|y_z=1,t_k) \\
&=\Pi_{k\in T_z} p( (l_x,l_y) - t_k | y_z=1 )
\end{align}
where $T_z$ is a set of joints that are informative about the bounding box center location of the object belonging to the class $z$. The algorithm to determine $T_z$ for each fashion item class $z$ will be described shortly. Each $p( (l_x,l_y) - t_k | y_z=1 )$ models the relative location of the bounding box center with respect to the $k$-th joint location. 

Intuitively, the locations of fashion items and those of body joints have strong correlations. For instance, the location of hat should be close to the location of head and thus, the distribution of their offset vector, $p( (l_x, l_y) - t_{\mathrm{Head}} | y_{\mathrm{Hat}} = 1 )$ should have a strong peak around $t_{\mathrm{Head}}$ and relatively easy to model. On the other hand, the location of left hand is less informative about the location of the hat and thus, $p( (l_x, l_y) - t_{\mathrm{Left hand}} | y_{\mathrm{Hat}} = 1 )$ typically have scattered and complex distribution which is difficult to model appropriately. Thus, it is beneficial to use for each fashion item only a subset of body joints that have strong correlations with the location of that item.

The relative location of the objects with respect to the joints can be most faithfully modeled as a multimodal distribution. For instance, bags, purses and wallets are typically carried on either left or right hand side of the body, thus generating multimodal distributions. To confirm this claim, In Figure \ref{fig:distribution}, we show a plot of $(l_x, l_y) - t_{\mathrm{Neck}}$ of ``Bag'' and a plot of $(l_x, l_y) - t_{\mathrm{Left Ankle}}$ of ``Left Shoe'' obtained from the dataset used in our experiments. As can be seen, $p( (l_x, l_y) - t_{\mathrm{Neck}} | y_{\mathrm{Bag}} = 1 )$ clearly follows a multimodal distribution while $p( (l_x, l_y) - t_{\mathrm{Left Ankle}} | y_{\mathrm{Left Shoe}} = 1 )$ has a unimodal distribution. Depending on the joint-item pair, it is necessary to automatically choose the number of modes.

To address the challenges raised above, we propose an algorithm to automatically identify the subset of body joints $T_z$ and learn a model. For each pair of a fashion item $z$ and a body joint $k$, we model $p( (l_x, l_y) - t_{k} | y_z = 1 )$ by a Gaussian mixture model (GMM) and estimate the parameters by the EM-algorithm. We determine the number of GMM components based on the Bayesian Information Criteria \cite{Kashyap1977,Schwarz1978} to balance the complexity of the model and fit to the data. To obtain $T_z$ for item $z$, we pick the top 2 joints whose associated GMM has larger likelihood. This way, for each item, body joints which have less scattered offsets are automatically chosen. The selected joints for each item will be shown in the next section.


{\renewcommand{\arraystretch}{1.2}%
\begin{table*}[tb]
\small
\centering
    \begin{tabular}{|c|c|c|c|c|c|c|c|c|c|c|c|}
    \hline
    \textbf{Methods} & \textbf{mAP} & \textbf{Bag} & \textbf{Belt} & \textbf{Glasses} & \textbf{Hat} & \textbf{Pants} & \textbf{Left Shoe} & \textbf{Right Shoe} & \textbf{Shorts} & \textbf{Skirt} & \textbf{Tights} \\ \hline
    \textbf{Full} & \textbf{31.1} & \textbf{22.5} & \textbf{14.2} &  \textbf{22.2} & \textbf{36.1} & \textbf{57.0} & \textbf{28.5} & \textbf{32.5} & \textbf{37.4} & \textbf{20.3} & \textbf{40.6} \\ \hhline{------------}
    \textbf{w/o geometric priors} & 22.9 & 19.4 & 6.0 & 13.0 & 28.9 & 37.2 & 20.2 & 23.1 & 34.7 & 15.2 & 31.7 \\ \hhline{------------}
    \textbf{w/o appearance} & 17.8 & 4.3 & 7.1 & 7.5 & 8.9 & 50.7 & 20.5 & 23.4 & 15.6 & 18.0 & 22.3 \\ \hline
    \end{tabular}
\caption{Average Precision of each method. ``Full'' achieves better mAP and APs for all the items than ``w/o geometric priors'' and ``w/o appearance''.}
\label{table:precision}
\end{table*}
}

{\renewcommand{\arraystretch}{1.2}%
\begin{table*}[ht]
\centering
	\begin{tabular}{|c|c|c|c|c|c|c|c|c|c|c|c|}
    \hline
\textbf{Bag} & \textbf{Belt} & \textbf{Glasses} & \textbf{Hat} & \textbf{Pants} & \textbf{Left shoe} & \textbf{Right shoe} & \textbf{Shorts} & \textbf{Skirt} & \textbf{Tights} & \textbf{Background} \\ \hline
    1,254 & 318 & 177 &  306 & 853 &  1,799 &  1,598 &  473 & 683 & 986  & 225,508 \\ \hline
    \end{tabular}
\caption{The number of training patches generated for each class with Selective Search~\cite{Uijlings2013}.}
\label{table:numtrain}
\end{table*}
}

{\renewcommand{\arraystretch}{1.2}%
\begin{table*}[ht]
\centering
    \begin{tabular}{|c|c|c|c|c|c|c|c|c|c|c|c|p{1.1cm}|}
    \hline
    \multirow{2}{1.2cm}{\centering \textbf{Precision (\%)}} & \multicolumn{11}{c|}{\textbf{Recall (\%)}} & \multirow{2}{1.2cm}{ \centering \textbf{Avg. \# of BBox}} \\
      \cline{2-12} & \textbf{Avg.} & \textbf{Bag} & \textbf{Belt} & \textbf{Glasses} & \textbf{Hat} & \textbf{Pants} & \textbf{L. Shoe} & \textbf{R. Shoe} & \textbf{Shorts} & \textbf{Skirt} & \textbf{Tights} & \\ \hline
    1.36 & 86.7 & 93.6 & 69.2 & 62.5 & 95.3 & 93.6 & 86.6 & 82.4 & 93.2 & 98.8 & 91.2 & 1073.4 \\ \hline    
    \end{tabular}
\caption{Precision, recall and the average number of generated bounding boxes per image. Note that it is important to have high recall and not necessarily precision so that we will not miss too many true objects. Precision is controlled later by the classification stage.}    
\label{table:bbrecall}
\end{table*}
}

\section{Experiments}
\label{sec:experiments}
\subsection{Dataset}
\label{sec:dataset}
To evaluate the proposed algorithm, we use the Fashionista Dataset which was introduced by~\cite{Yamaguchi2012} for pixel-level clothing segmentation. Each image in this dataset is fully annotated at pixel level, \textit{i.e.} a class label is assigned to each pixel. In addition to pixel-level annotations, each image is tagged with fashion items presented in the images. In~\cite{Yamaguchi2013}, another dataset called Paper Doll Dataset including 339,797 tagged images is introduced and utilized to boost performance on the Fashionista Dataset. Our method does not use either associated tags or the Paper Doll Dataset. We use the predefined training and testing split for the evaluation (456 images for training and 229 images for testing) and take out 20\% of the training set as the validation set for the parameter tuning.

In the Fashionista Dataset, there are 56 classes including 53 fashion item classes and three additional non-fashion item classes (hair, skin and background.) We first remove some classes that do not appear often in the images and those whose average pixel size is too small to detect. 
We then merge some classes which look very similar. For instance, there are ``bag'', ``Purse'' and ``Wallet'' classes but the distinction between those classes are visually vague, thus we merge those three classes into a single ''Bag'' class. We also discard all the classes related to footwear such as ``sandal'' and ``heel' and instead add ``left shoe'' and ``right shoe'' classes which include all types of footwear. It is intended that, if needed by a specific application, a sophisticated fine-grained classification method can be applied as a post-processing step once we detect the items. Eventually, we obtain 10 new classes where the occurrence of each class is large enough to train the detector and the appearance of items in the same class is similar. The complete definition of the new 10 classes and some statistics are shown in Table~\ref{table:classDefinition}.

We create ground-truth bounding boxes based on pixel-level annotations under the new definition of classes. For classes other than ``Left shoe'' and ``Right shoe'', we define a ground-truth bounding box as the one that tightly surrounds the region having the corresponding class label. For ``Left shoe'' and ``Right shoe'' classes, since there is no distinction between right and left shoes in the original pixel-level annotations, this automatic procedure cannot be applied. Thus, we manually annotate bounding boxes for ``Right shoe'' and ``Left shoe'' classes. These bounding box annotations will be made available to facilitate further research on fashion apparel detection.

Our framework is general in the choice of pose estimators. In this work, we use pose estimation results provided in the Fashionista Dataset, which is based on \cite{Yang2011}. There are 14 key joints namely head, neck, left/right shoulder, left/right elbow, left/right wrist, left/right hip, left/right knee and left/right foot. 

In Table~\ref{table:classDefinition}, we show the first and second key body joints that are selected by the proposed algorithm. Interestingly, for ``Pants'', ``Shorts'' and ``Skirt'', left hip and right hip are selected but for ``Tights'', left knee and right knee are selected instead.


\subsection{Detector Training}\label{sec:detectorTraining}
We create image patches for detector training by cropping the training images based on the corresponding ground-truth bounding box. Before cropping, we enlarge the bounding boxes by a scale factor of 1.8 to include the surrounding regions, thus providing contextual information. Note that we intentionally make the contextual regions larger than \cite{Girshick2014} as contextual information would be more important when detecting small objects like fashion items we consider in this work. The cropped image patches are then resized to the size of the first layer of CaffeNet (227 by 227 pixels). To increase the number of training patches, we run the object proposal algorithm on the training images and for each generated bounding box, we compute the intersection over union (IoU) with the ground-truth bounding boxes. If the IoU is larger than 0.5 for a particular class, we use the patch as an additional training instance for that class. If IoU is smaller than 0.1 with ground-truth bounding boxes of all the classes, we use it as a training instance for a background class. We also obtain training patches for the background class by including image patches from ground-truth bounding boxes of the classes which we do not include in our new 10 classes.

The number of training patches for each class obtained are shown in Table~\ref{table:numtrain}. From the obtained training patches, we train a set of linear SVMs, each of which is trained by using instances in a particular class as positive samples and all instances in the remaining classes as negative samples. The parameters of SVMs are determined from the validation set.


\subsection{Baseline Methods}
Since fashion apparel detection has not been previously addressed, there is no existing work proposed specifically for this task. Thus, we convert the pixel-level segmentation results of~\cite{Yamaguchi2012} and~\cite{Yamaguchi2013} to bounding boxes and use their performance as baselines. To obtain bounding boxes from  segmentation results, we use the same procedure we use to generate ground-truth bounding boxes from the ground-truth pixel-level annotations. Note that we exclude ``Left shoe'' and ``Right shoe'' from the comparison since in their results, there is no distinction between left and right shoes.  


\subsection{Results}
We first evaluate the performance of the object proposal methods in terms of precision and recall. Here, precision is defined as the number of object proposals which match the ground-truth bounding boxes regardless of class, divided by the total number of object proposals. Specifically, we consider each object proposal as correct if $\mathrm{IoU} \geq 0.5$ for at least one ground-truth bounding box. We compute recall for each class by the number of ground-truth bounding boxes which have at least one corresponding object proposal, divided by the total number of ground-truth bounding boxes. 

In Table \ref{table:bbrecall}, we show the precision, recall and the average number of object proposals per image. We tune the parameters of both object proposal algorithms to retain high recall so that it will not miss too many true objects. Although it results in the low precision, false positives are reduced in the subsequent classification stage.


We evaluate the performance of the detection methods using the average precision (AP) computed from the Precision-Recall curves. In Table~\ref{table:precision}, we report the performance of the proposed framework with three different settings, ``Full'' represents our complete method using both geometric priors and appearance-based posterior, ``w/o geometric prior'' represents a method which excludes the geometric priors from ``Full'' and ``w/o appearance'' is a method which excludes appearance-based posterior from ``Full''.

From the comparison between ``Full'' and ``w/o geometric prior'', it is clear that incorporating geometric priors significantly improves the performance (35.8\% improvement for mAP). This result indicates the effectiveness of the geometric priors in the fashion item detection task. 





\begin{figure*}[tp]  
\centering
  \subcaptionbox{Bag}{\includegraphics[width=0.24\linewidth]{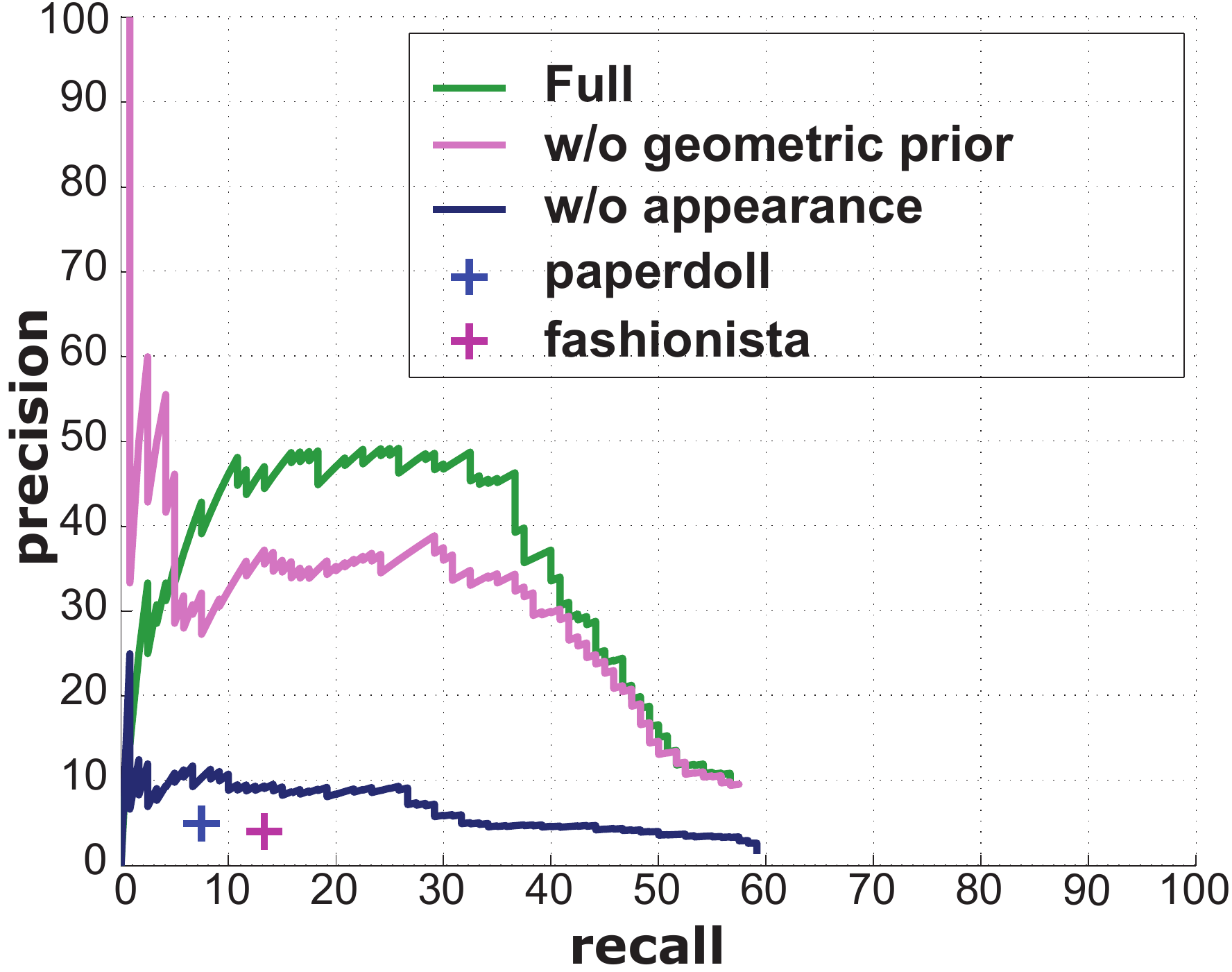}}
  \subcaptionbox{Belt}{\includegraphics[width=0.24\linewidth]{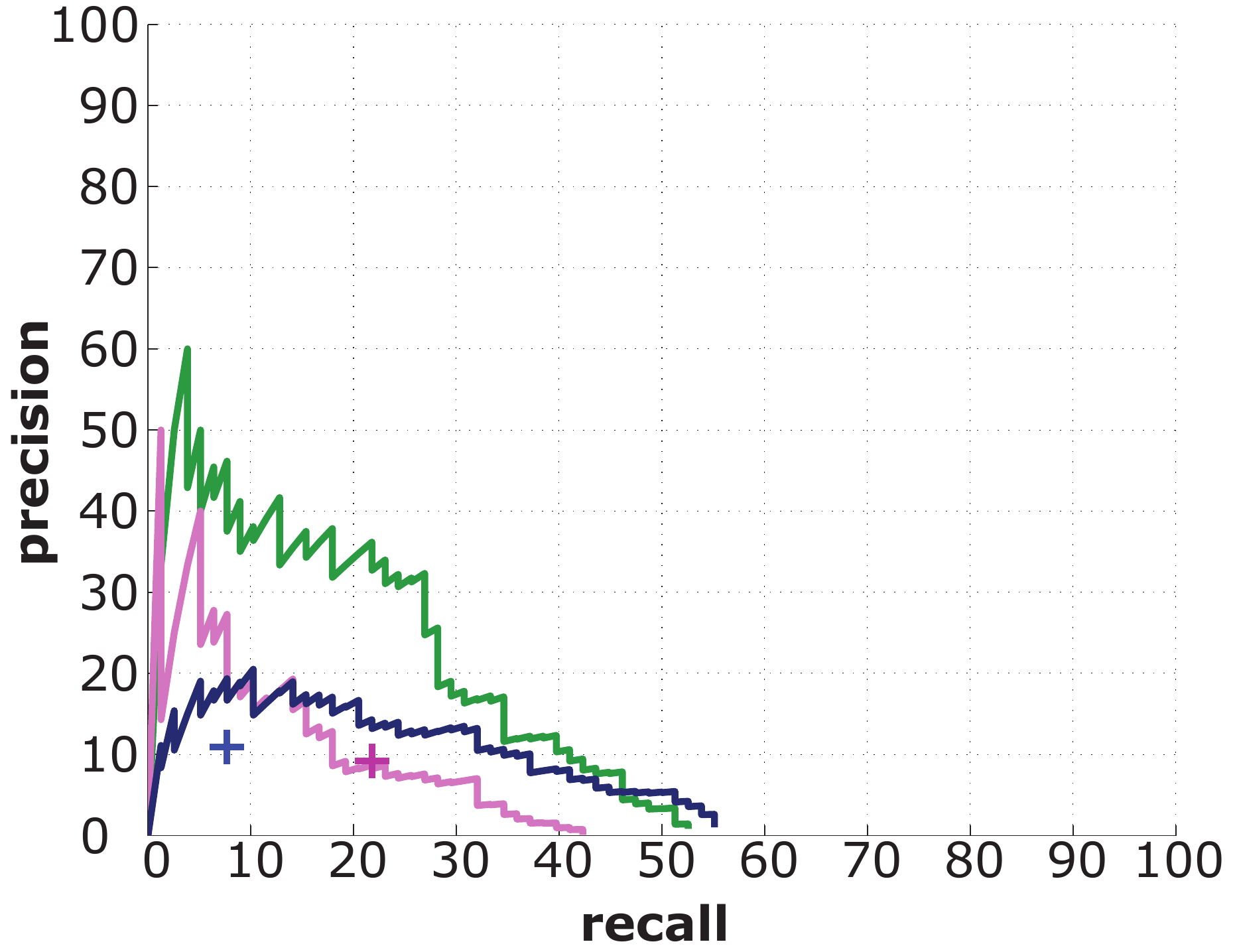}}
  \subcaptionbox{Glasses}{\includegraphics[width=0.24\linewidth]{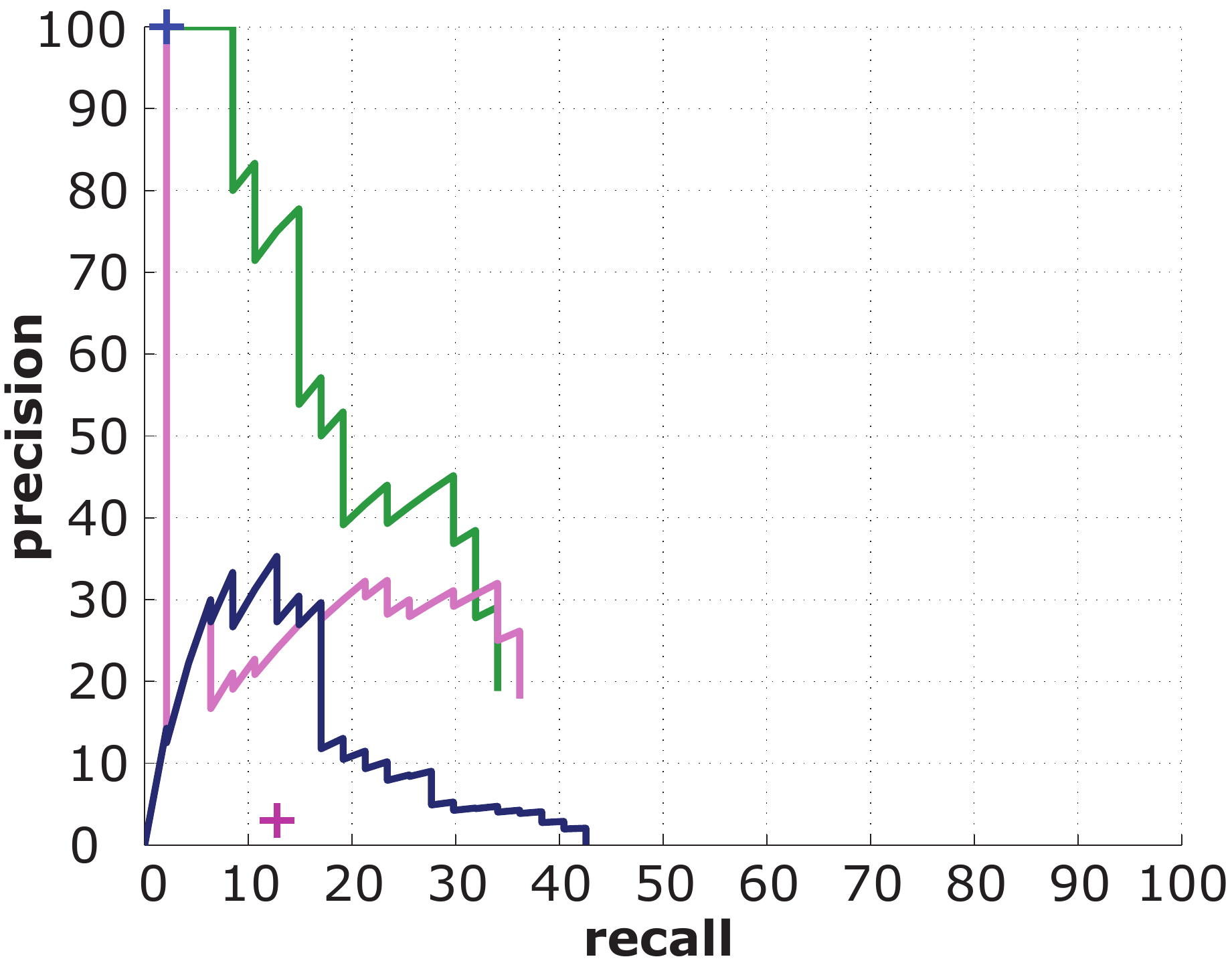}}
  \subcaptionbox{Hat}{\includegraphics[width=0.24\linewidth]{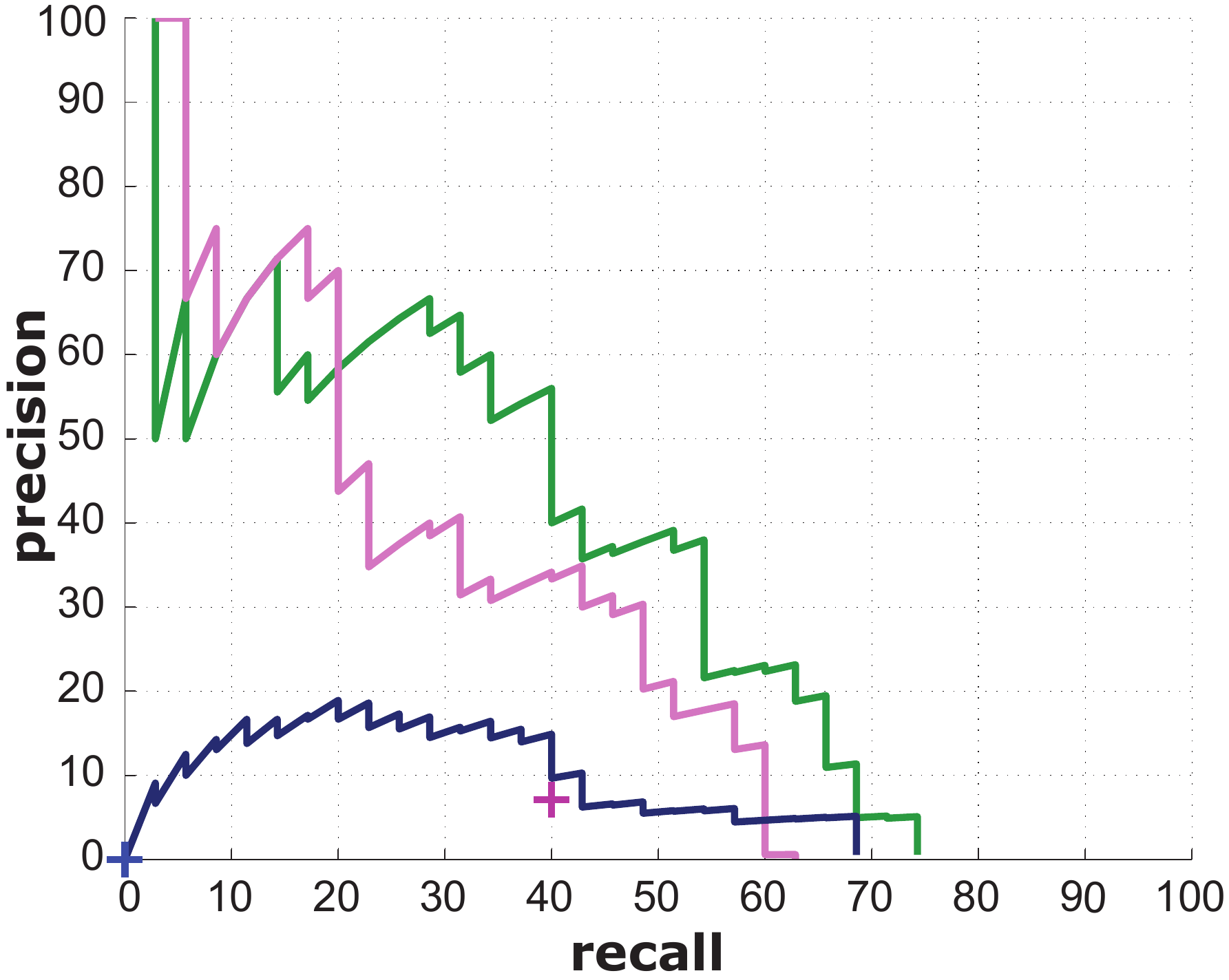}}
  \subcaptionbox{Pants}{\includegraphics[width=0.24\linewidth]{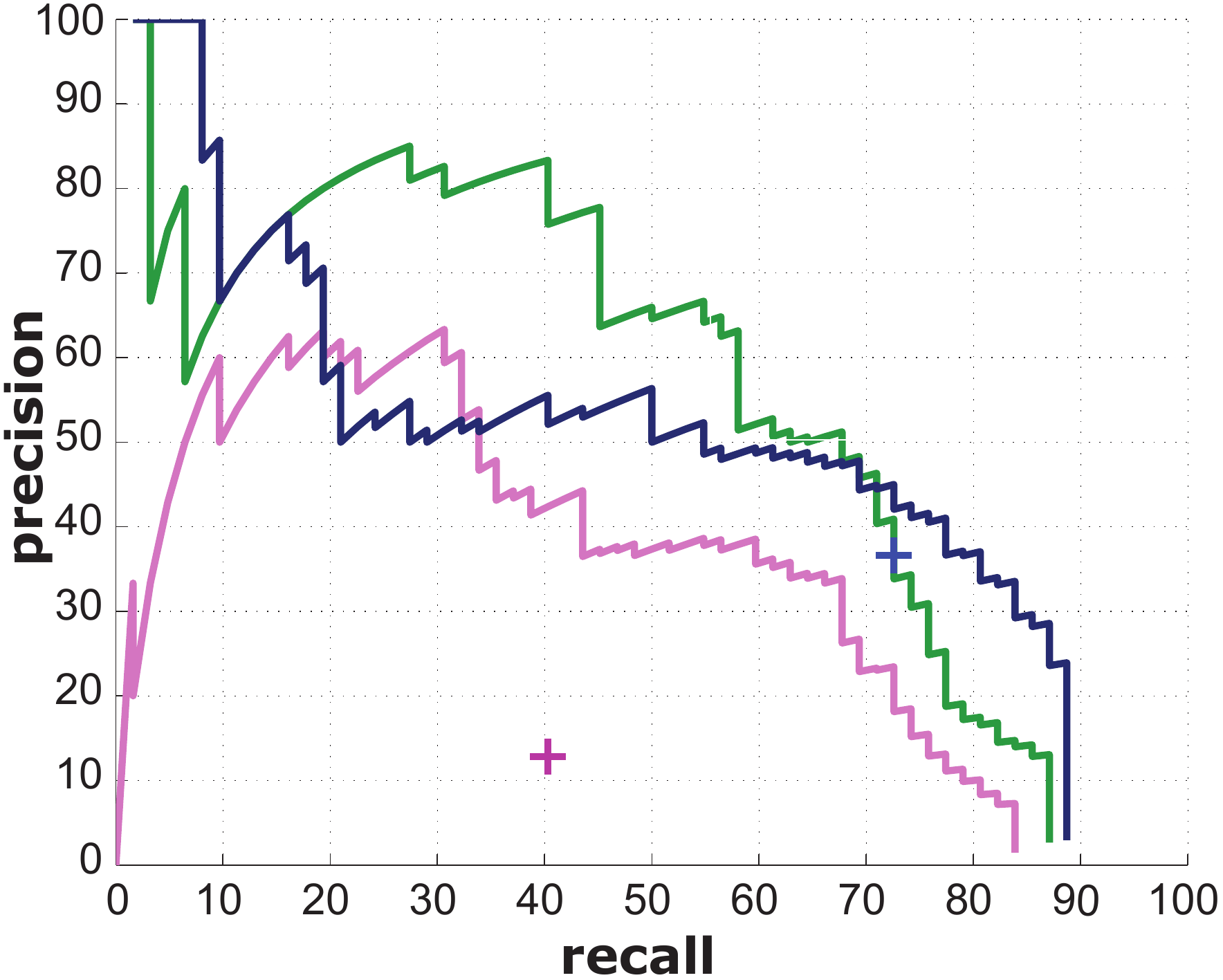}}
  \subcaptionbox{Left shoe}{\includegraphics[width=0.24\linewidth]{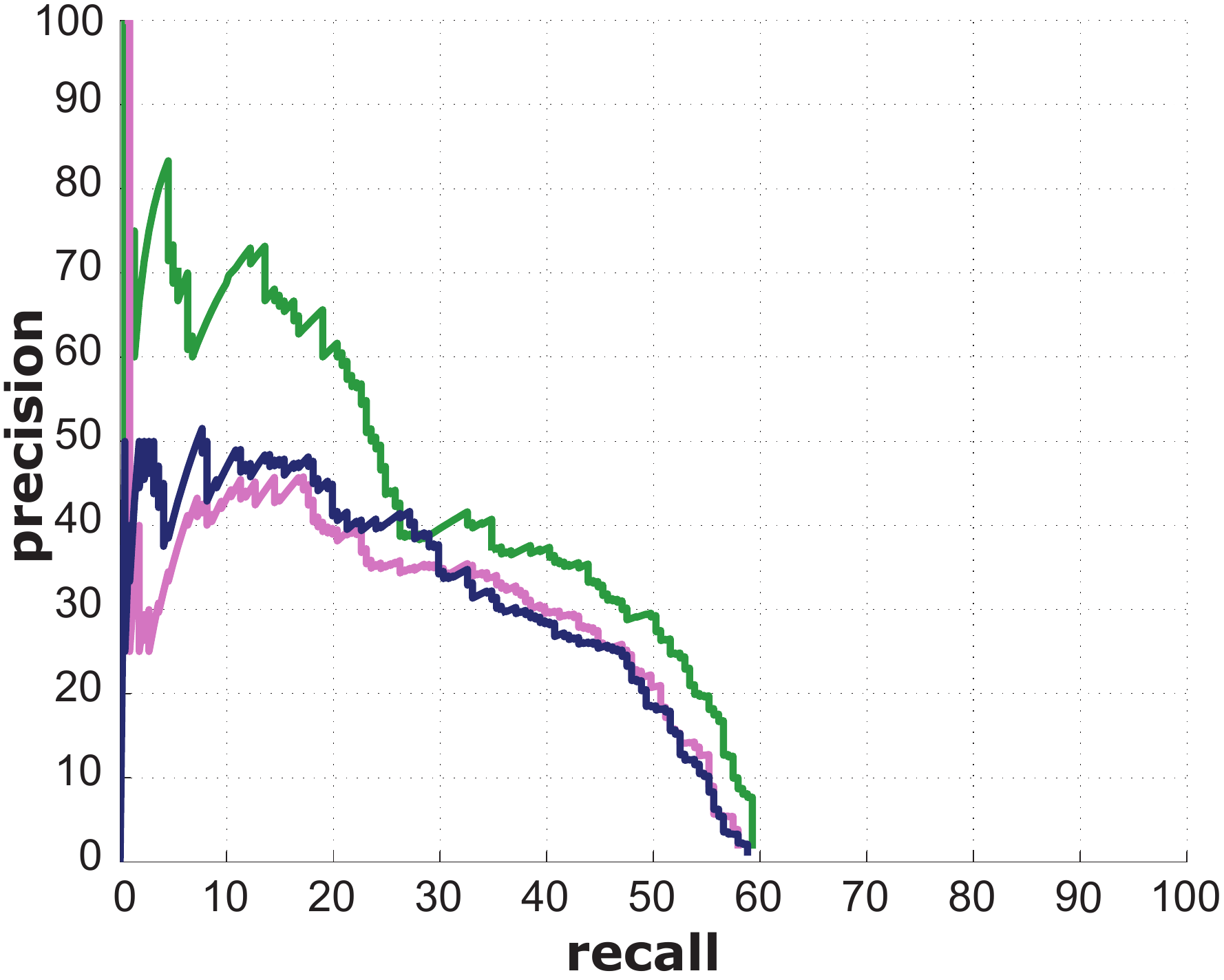}}
  \subcaptionbox{Right shoe}{\includegraphics[width=0.24\linewidth]{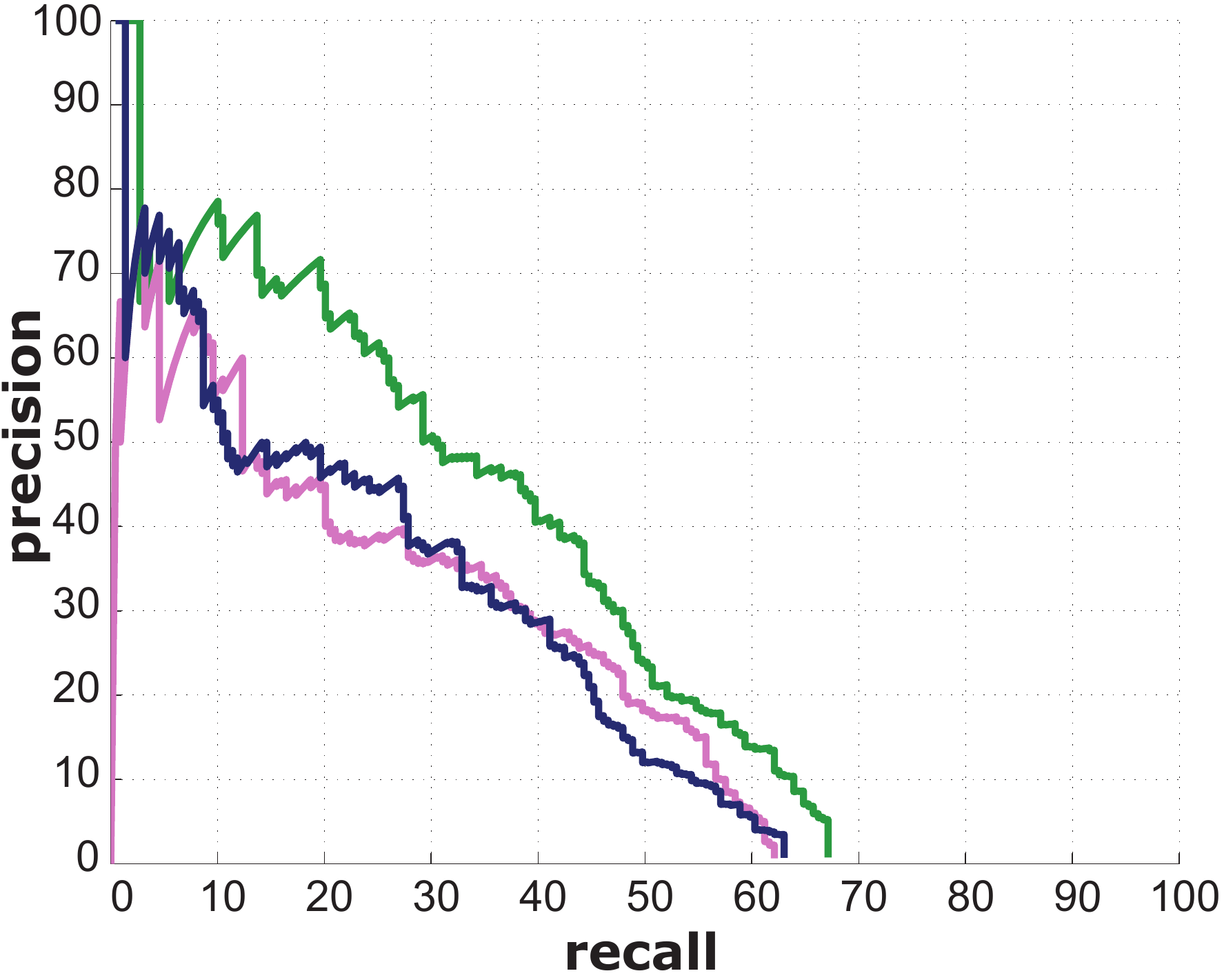}} 
  \subcaptionbox{Shorts}{\includegraphics[width=0.24\linewidth]{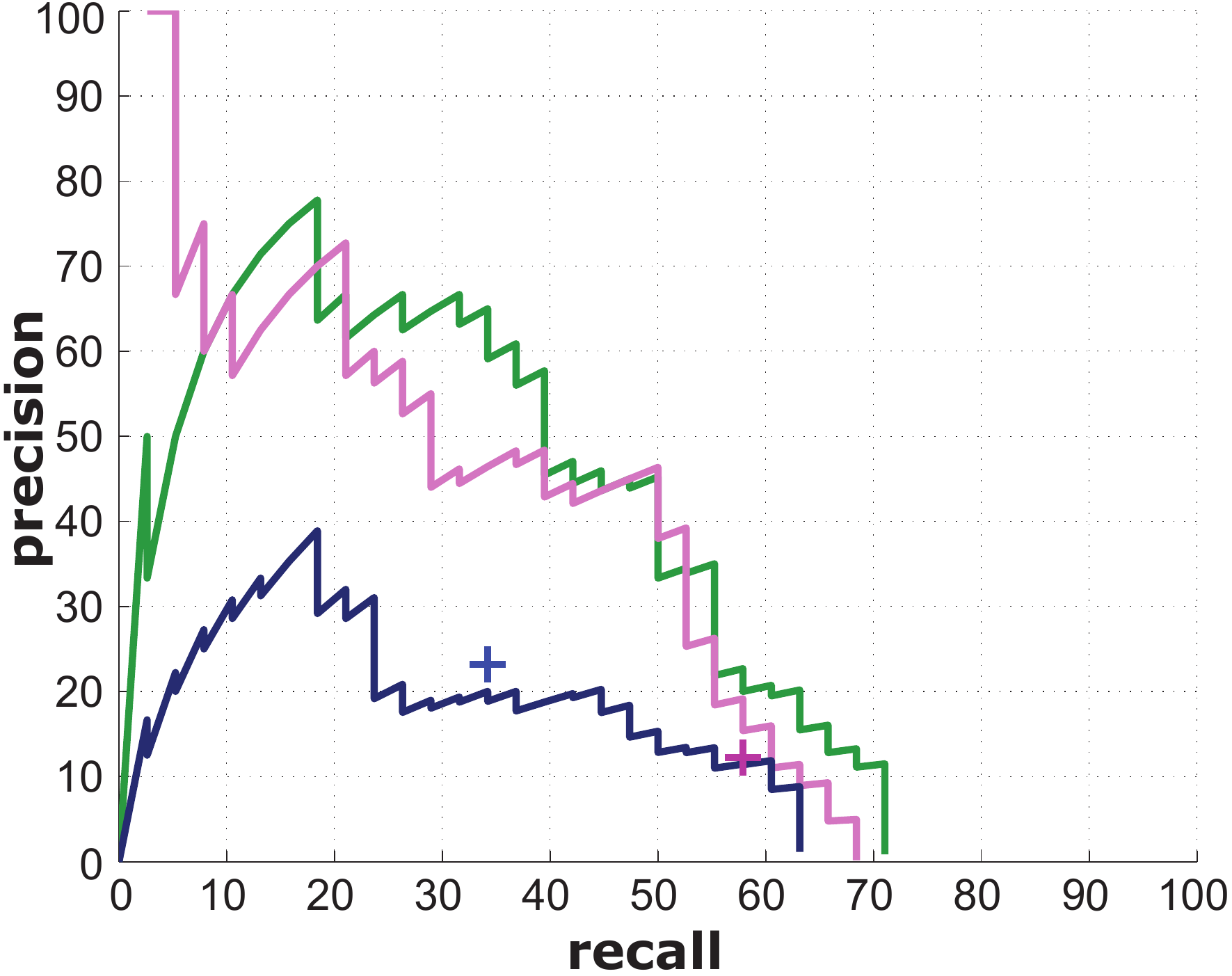}} 
  \subcaptionbox{Skirt}{\includegraphics[width=0.24\linewidth]{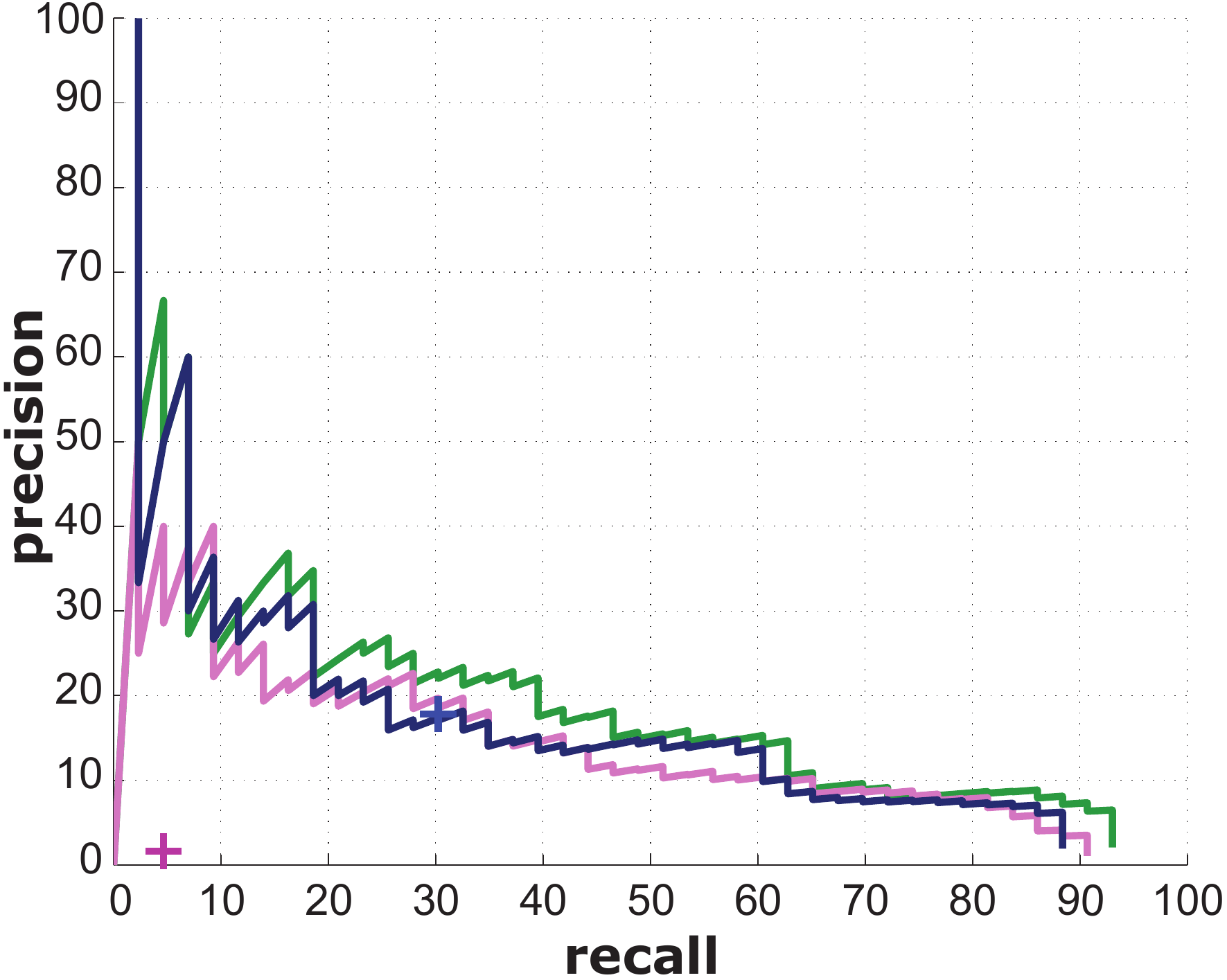}}
  \subcaptionbox{Tights}{\includegraphics[width=0.24\linewidth]{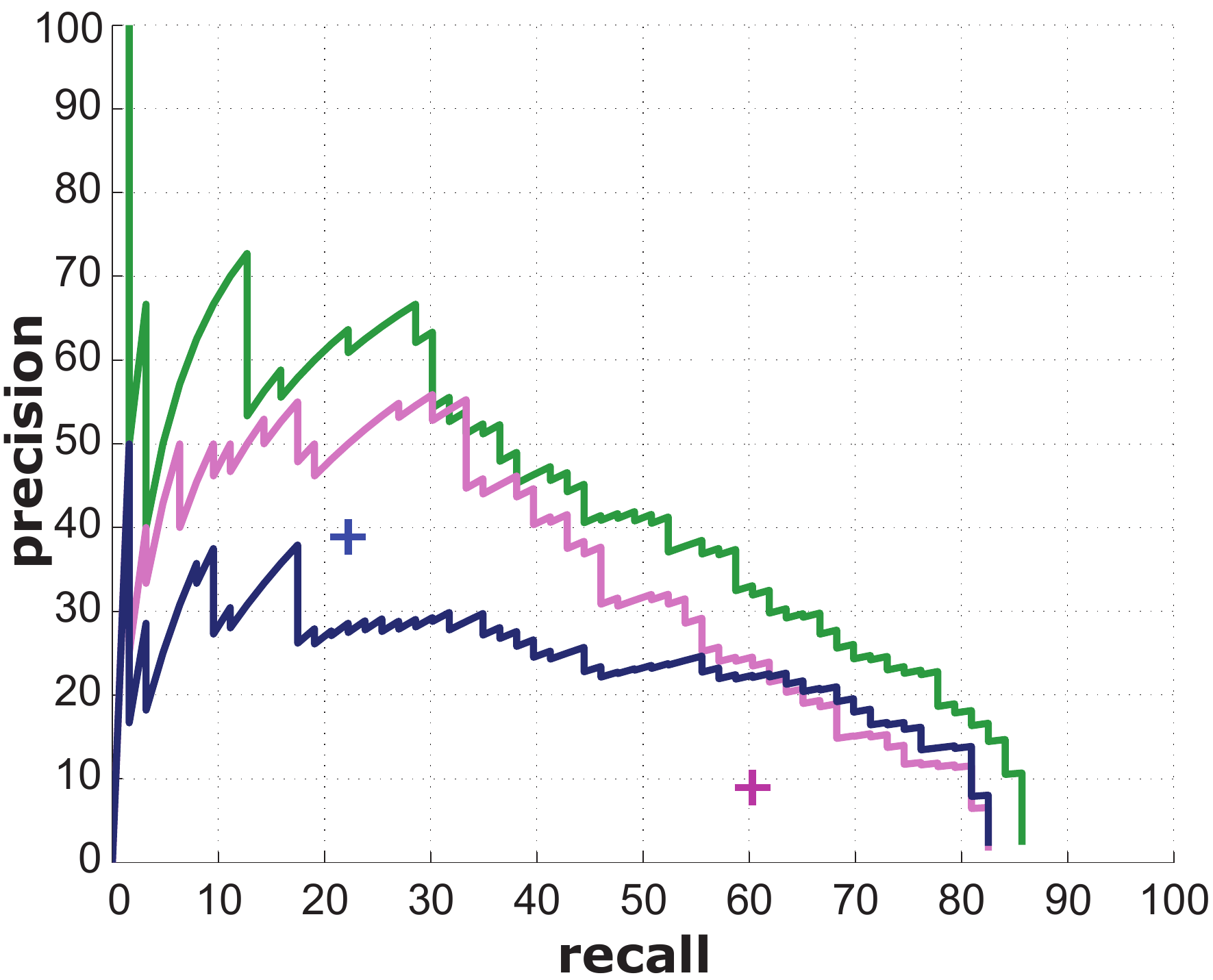}}
\caption{Precision-Recall curves for each fashion category. Our full method outperforms the baseline method (shown by cross) with a large margin (sometimes up to 10 times in precision for the same recall), except for ``Pants''. Note that we do not have results from the baseline methods for ``Left shoe'' and ``Right shoe'' as they are newly defined in this work.}
 \label{fig:PR-curves_SS}
\end{figure*}

In Figure \ref{fig:PR-curves_SS} we show precision-recall curves of the proposed methods with various settings as well as precision-recall points of the baseline methods. In the figures, ``paperdoll'' refers to the results of \cite{Yamaguchi2013} and ``fashionista'' refers to \cite{Yamaguchi2012}. Except for ``Pants'', our complete method outperforms the baselines with a large margin. Note that ``paperdoll'' \cite{Yamaguchi2013} uses the large database of tagged fashion images as additional training data.

\begin{figure}[htb]
  \centering
    \includegraphics[width=3.3in]{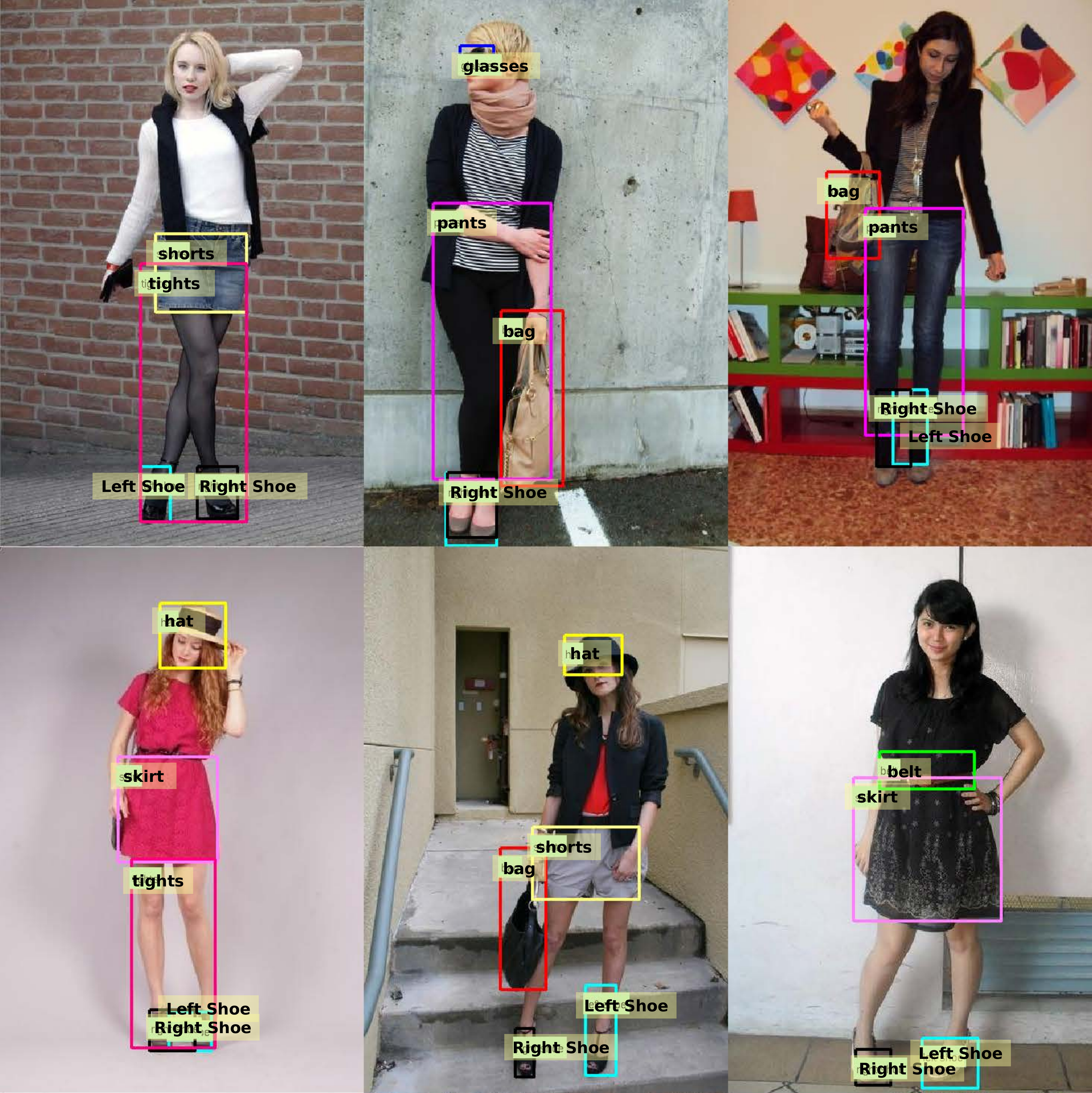}
  \caption{Example detection results obtained by the proposed method. Note that we overlaid text labels manually to improve legibility.}
  \label{fig:visualresults}
\end{figure}

\begin{figure*}[htp]
  \centering
    \includegraphics[width=6.0in]{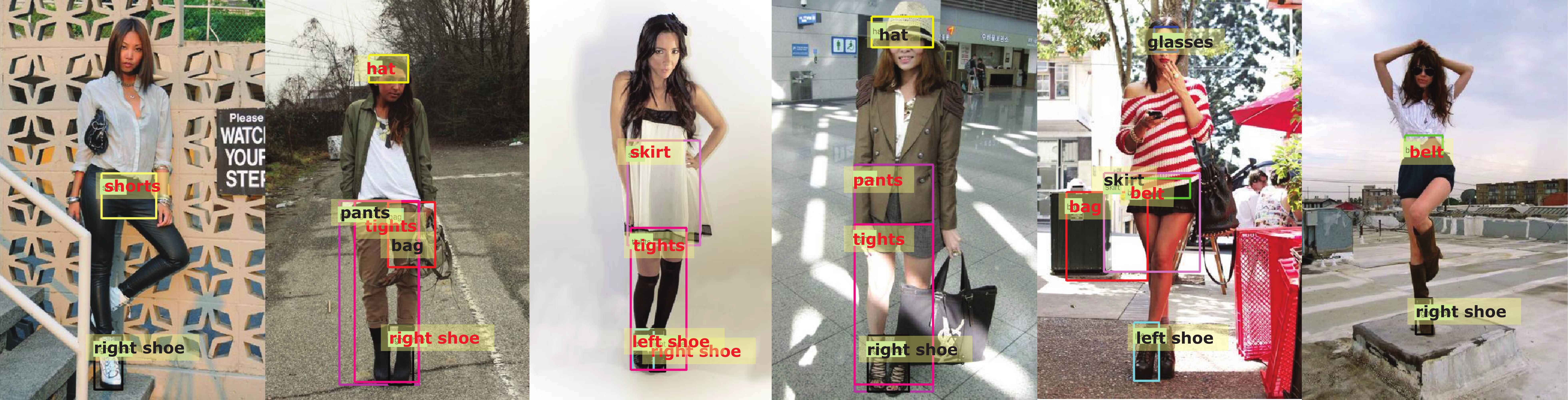}
  \caption{Examples of failed detection results obtained by the proposed method. Note that we overlaid text labels manually to improve legibility. Incorrect labels are shown in red.}
  \label{fig:bad}
\end{figure*}

In Figure~\ref{fig:visualresults}, we show some qualitative results. Figure~\ref{fig:bad} shows sample images where our approach makes mistakes. We argue that fashion apparel detection has its own unique challenges. First of all, even with our new fashion item classes, some fashion items are visually very similar to each other. For example, ``Tights'' and ``Pants'' can look very similar since both items can have a variety of colors. The only distinguishable cue might be how tight it is, which is quite challenging to capture. Another example is ``Skirt'' and bottom half of a dress. Both items have extremely similar appearance. The only difference is that a dress is a piece of cloth which covers both upper body and lower body and this difference is difficult to detect. Furthermore, ``Belt'' and ``Glasses'' are difficult to detect as they are usually very small. 



\section{Conclusion}
\label{sec:conclusions}
In this work, we reformulate fashion apparel parsing, traditionally treated as a semantic segmentation task, as an object detection task and propose a probabilistic model which incorporates state-of-the-art object detectors with various geometric priors of the object classes. Since the locations of fashion items are strongly correlated with the pose of a person, we propose a pose-dependent prior model which can automatically select the most informative joints for each fashion item and learn the distributions from the data. Through experimental evaluations, we observe the effectiveness of the proposed priors for fashion apparel detection. 

{
\bibliographystyle{ieee}
\bibliography{library}
}

\end{document}